\documentclass[letterpaper, 10 pt, conference]{ieeeconf}  

\IEEEoverridecommandlockouts                              

\UseRawInputEncoding
\usepackage{graphics} 
\usepackage{graphicx}
\usepackage{subfigure}
\usepackage{amsmath} 

\title{\LARGE \bf
An Explicit Method for Fast Monocular Depth Recovery in Corridor Environments 
}

\author{Yehao liu$^{1}$, Ruoyan Xia$^{1}$, Xiaosu Xu$^{1}$,Zijian Wang$^{1}$,Yiqing Yao$^{1}$ \IEEEmembership{Member, IEEE}, and Mingze Fan$^{1}$
\thanks{$^{1}$Albert Author is with 
	Key Laboratory of Micro-Inertial Instruments and Advanced Navigation Technology, Ministry of Education,
	School of Instrument Science and Engineering, Southeast University, Nanjing 210096, China
        }
}

\begin{document}

\maketitle
\thispagestyle{empty}
\pagestyle{empty}

\begin{abstract}

Monocular cameras are extensively employed in indoor robotics, but their performance is limited in visual odometry, depth estimation, and related applications due to the absence of scale information.Depth estimation refers to the process of estimating a dense depth map from the corresponding input image, existing researchers mostly address this issue through deep learning-based approaches, yet their inference speed is slow, leading to poor real-time capabilities. To tackle this challenge, we propose an explicit method for rapid monocular depth recovery specifically designed for corridor environments, leveraging the principles of nonlinear optimization. We adopt the virtual camera assumption to make full use of the prior geometric features of the scene. The depth estimation problem is transformed into an optimization problem by minimizing the geometric residual. Furthermore, a novel depth plane construction technique is introduced to categorize spatial points based on their possible depths, facilitating swift depth estimation in enclosed structural scenarios, such as corridors. We also propose a new corridor dataset, named	Corr\_EH\_z, which contains images as captured by the UGV camera of a variety of corridors. An exhaustive set of experiments in different corridors reveal the efficacy of the proposed algorithm.

\end{abstract}

\section{INTRODUCTION}

Monocular cameras play a pivotal role in indoor robotics\cite{khan2020deep,dong2022towards,ming2021deep};
 nevertheless, their performance is constrained in certain application fields, such as visual odometry and 3D object detection, due to the absence of scale information \cite{khan2020deep,laga2020survey,vyas2022outdoor,mur2017orb}. To address this limitation and obtain scale information from images, researchers often employ supplementary auxiliary methods. While RGBD cameras offer the advantage of obtaining relatively highprecision scene depth images, their resolution remains comparably low, rendering them susceptible to the influence of deep black objects, translucent materials, specular reflections, and parallax effects, thereby leading to reduced accuracy. Conversely, stereo cameras can calculate pixel depth through triangulation; however, this advantage comes at the expense of increased computational overhead, and their distance estimation is subject to limitations imposed by the baseline length. 

Depth estimation refers to the process of estimating a dense depth map from the corresponding input image\cite{dong2022towards}. Monocular depth estimation holds significant research value \cite{vyas2022outdoor,ming2021deep}. When combined with object detection, it can achieve the effect of 3D reconstruction of general Lidar detection\cite{park2021pseudo}. Furthermore, through integration with semantic segmentation, the approach can be extended from 2D to 3D, allowing the acquisition of both semantic and depth information for pixels.

Monocular depth estimation is an ill-posed problem that requires the introduction of sufficient prior information for its resolution. Monocular depth estimation methods can be categorized into structure from motion (SFM) based methods\cite{ha2016high,javidnia2017accurate,wang2018learning,yang2018deep}, hand-crafted feature based methods, and deep learning-based methods\cite{eigen2014depth,wang2020depthnet,bhat2021adabins,fu2018deep}. Each approach explores different strategies to address the challenge of recovering depth from a single camera input. 
With the rapid advancement of deep neural networks, monocular depth estimation based on deep learning has attracted considerable interest and demonstrated remarkable accuracy\cite{zhao2020monocular,vyas2022outdoor}. The impressive performance of deep learning methods relies on thorough training on extensive datasets, and the accuracy heavily hinges on the quality of precisely annotated data. Acquiring highquality data for depth/parallax reconstruction involves substantial time and labor costs \cite{vyas2022outdoor}. Furthermore, deep learning-based methods exhibit limited generalization capacity in depth estimation due to the influence of image size and scene characteristics present in the training data\cite{laga2020survey}. Additionally, the majority of deep learning approaches demonstrate slow inference speeds and insufficient realtime performance.

SfM-based methods perform 3D scene reconstruction using multiple image sequences from different perspectives. They extract feature points from the images for feature matching, estimate camera motion and 3D positions of pixels, and construct sparse depth maps by assembling point cloud information of 3D space points\cite{ha2016high,javidnia2017accurate,wang2018learning,yang2018deep}. However, SfM-based methods require matching alignment between multiple frames with continuous motion, and their accuracy is highly dependent on the results of inter-frame registration, thereby limiting their application in certain scenarios.

Long corridors/hallways are characteristic of challenging scenarios with limited texture features, and a high degree of similarity between frames hinders reliable inter-frame alignment. Thus, SfM-based methods are susceptible to failure and reduced accuracy in such degraded scenes. Nonetheless, long corridors/hallways manifest strong structured characteristics, encompassing abundant geometric information, such as parallel walls on both sides and maintaining parallel lines at the junction of the floor and walls. By fully leveraging these structured features, Inverse Projective IPM (Inverse Projective IPM)\cite{jeong2016adaptive,lin2023adaptive,wang2015inverse,yang2016fast} can be employed to derive scale information for these characteristics, enabling depth plane construction and subsequent scene depth recovery.

In this context, we propose a novel display method for rapid monocular depth estimation. Our approach differs from existing methods as it eliminates the need for inter-frame matching assistance and avoids extensive training on large datasets, thereby saving on model training and transfer costs. By leveraging the virtual camera assumption and minimizing geometric residuals, we transform the depth estimation problem into an optimization task. Furthermore, we introduce a depth plane construction method, which categorizes spatial points based on their possible depths, enabling fast depth estimation in enclosed structural scenes such as long corridors/hallways. Our proposed method achieves state-of-the-art depth estimation accuracy in long corridor/hallway scenarios while significantly accelerating the depth recovery process. Moreover, it can achieve realtime monocular depth recovery on mid-to-low-performance processors.

\section{Related Work}

\subsection{Explicit Method for Depth Estimation}


The explicit method for depth estimation refers to an approach in which the entire process of depth estimation, from feature extraction and feature transformation to the output of prediction results, can be explained using mathematical formulas. This method is commonly employed in depth estimation techniques based on SFM. On the other hand, implicit methods achieve the same task through techniques such as convolutional neural networks (CNNs), where the processes of feature extraction, feature space transformation, and depth prediction are encapsulated within an end-to-end deep network model.

The SFM algorithm receives input image sequences taken from different viewing angles and first extracts features such as Harris, SIFT or SURF from all images. Feature matching is then performed to estimate the 3D coordinates of the features and generate a point cloud that can be converted into a depth map. 
In 2014, Prakash et al. \cite{prakash2014sparse}proposed a sparse depth estimation method based on SFM. Based on monocular image sequences from 5 to 8 different perspectives, the method used a multi-scale fast detector for feature detection and 3D position solution based on geometric view relations to obtain a sparse depth map. 

In 2016, Ha et al. \cite{ha2016high}proposed a Structure From Small Motion (SFM) recovery method, which uses planar scanning technology to estimate depth maps. By using Harris corner detection and optical flow tracking method to solve the 3D position of the feature points, a relatively dense depth map can be obtained, but this algorithm cannot run in real time in terms of speed. 

In recent years, researchers have attempted to combine the strengths of explicit and implicit methods. In 2022, Zhong et al. \cite{zhong2022snake} introduced a method that simultaneously conducts implicit reconstruction and extracts 3D feature points, while others usually use explicit method to get 3D points. It replaces manual feature extraction with an implicit description for 3D keypoint detection. 

In 2023, Wu et al. \cite{wu2023heightformer} demonstrated the equivalence of depth and height in the 2D-to-3D mapping transformation and proposed an explicit height description method applied to deep network models for transforming Bird's Eye View (BEV) space.

\subsection{Real-Time Monocular Depth Estimation}

The overall development trend of monocular depth estimation is to push the increase of accuracy using extremely deep
CNNs or by designing a complex network architecture, which
are computationally expensive for current mobile computational
devices which have limited memory and computational
capability. Therefore, it is difficult for these networks to
be deployed in small sized robots which depend on mobile
computational devices. Under this context, researchers have
begun to develop real-time monocular depth estiamtion methods\cite{dong2022towards}.

In 2018,Poggi et al. \cite{poggi2018towards} stack a simple encoder and multiple small decoders working in a pyramidal structure, which is capable to quickly infer an accurate depth map on a CPU, even of an embedded system, using a pyramid of features extracted from a single input image.The network was trained in an unsupervised manner casting depth estimation as an image reconstruction problem. The designed network only has 1.9M parameters and requires.
0.12s to produce a depth map on a i7-6700K CPU, which isclose to a real-time speed.

In 2019, Wofk et al. \cite{wofk2019fastdepth} develop a lightweight encoderdecoder network for monocular depth estimation.A low latency, high throughput, high accuracy depth estimation algorithm running on embedded systems was designed. In addition,a network pruning algorithm is applied to further reduce the amount of parameters, which enables real-time depth estimation on embedded platforms with an Nvidia-TX2 GPU.

In 2020, Wang et al. \cite{wang2020depthnet} design a highly compact network named
DepthNet Nano. DepthNet Nano applies densely connected
projection batchnorm expansion projection (PBEP) modules
to reduce network architecture and computation complexity
while maintaining the representative ability.

In 2020, Liu et al. \cite{liu2020mininet} introduce a lightweight
model (named MiniNet) trained on monocular video
sequences for unsupervised depth estimation. The core part of
MiniNet is DepthNet, which iteratively utilizes the recurrent
module-based encoder to extract multi-scale feature maps. The
obtained feature maps are passed to the decoder to generate
multi-scale disparity maps. MiniNet achieves real-time speed
about 54fps with 640 × 192 sized images on a single Nvidia
1080Ti GPU.

However, the accuracy of above is inferior to
state-of-the-art methods. Therefore, developing real-time monocular depth estimation
network is assumed to achieve the trade-off between accuracy
and efficiency.

\subsection{Corridor Environments Perception and Localization}
Long corridor is a typical degraded scene with a lack of texture features, which brings new challenges to visual perception and localization tasks\cite{padhy2018deep}. It is necessary to understand the characteristics of this scene, so as to make full use of prior features and achieve high-precision perception and localization. However, long corridor scenes are inevitably faced by mobile robots, in recent years, some researchers have begun to focus on solving this problem.

In 2021, Padhy et al.\cite{padhy2021localization} introduce a localization method of Unmanned Aerial Vehicles(UAV) in Corridor Environments, a Deep Neural Network(DNN) was trained to understand corridor environmental information, andpredict the position of the UAV as either on the left or center or right side of the corridor.Depending upon the divergence of the UAV with respect to an imaginary central line, known as the central bisector line (CBL) of the corridor, a suitable command is generated to bring the UAV to the center, making UAV fly safely in Corridors.

In 2023, Ge et al.\cite{ge2023visual} proposed a visual-feature-assisted localization methods in long corridor environments.A novel coarse-to-fine paradigm was presented that uses visual features to assist mobile robot localization in long corridors.Sufficient keyframes are obtained through the camera, and a visual camera map was created while the grid map built by the laser-based SLAM method with a low accuacy in corridors, and the mobile robot captures images in a proper perspective according to the moving strategy and matches them with the image map to achieve a coarse localization.

\section{Materials and Methods}
\subsection{System Overview}
%

\begin{figure}[ht]
	\centering
	\includegraphics[width=6 cm]{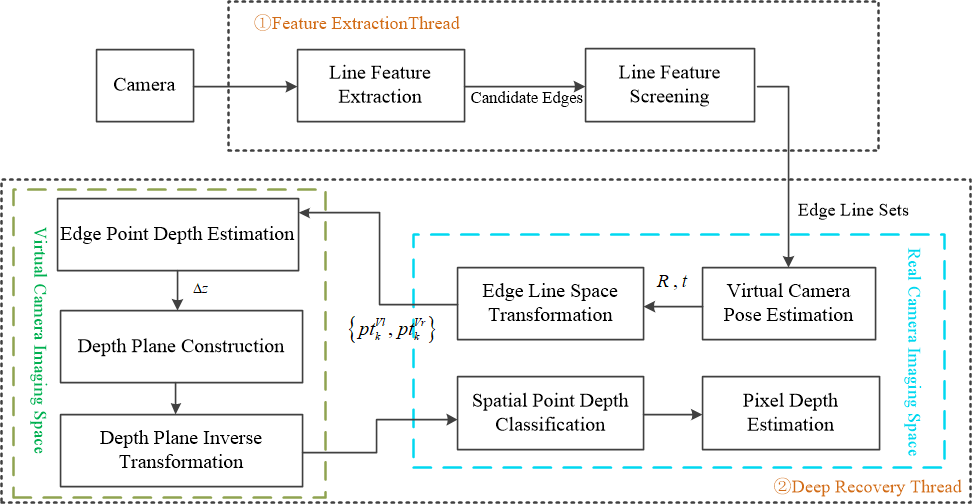}
	\caption{System framework.There are two threads.Thread 1 main function is to extract ground edges combined along the line. Thread 2 mainly completes the depth estimation of the scene.\label{fig1}}
\end{figure}   

The proposed method consists of two main threads: the edge extraction thread ,which main function is to extract ground edges line Sets, and the depth recovery thread, witch mainly completes the depth estimation of the scene.

As shown in Figure$~\ref{fig1}$. Images acquired from the visual sensor are first input to the feature extraction thread. In this thread, line feature information is extracted from the scene using Hough transform-based line feature detection. The line features of the sense are then filtered based on the distribution of line segment angles, leading to the construction of a set of ground edge lines. The information about the edge line set is then sent to the depth recovery thread.

In the depth recovery thread, the edge line set is first projected into the virtual camera imaging space. The current camera-to-virtual-camera pose transformation is estimated by minimizing symmetry geometric residuals. Subsequently, the edge line set is transformed to the virtual camera imaging plane using the computed transfor-mation matrix. Distance geometric residuals are then constructed, and a non-linear optimization process is iteratively performed to estimate the camera's pitch angle and the depths of the edge points.Based on the estimated depths of the edge points, depth planes are constructed and transformed back to the original image plane. Pixel points in the original image plane are classified based on the depth information, and finally, depth estimation values for the pixel points are obtained through approximate inter-polation.

\subsection{Edge Extration Thread}

Thread 1 primarily engages in the extraction of structured scene features through ground edge detection, thereby furnishing the depth recovery thread with essential prior information about the scene. In this study, it is assumed that the width of the long corridor/hallway remains constant, leading to the representation of ground and side wall edges as two straight lines within the image.

\begin{figure}[ht]
	\centering
	\subfigure[\label{fig:a}]{
		\includegraphics[width=3cm]{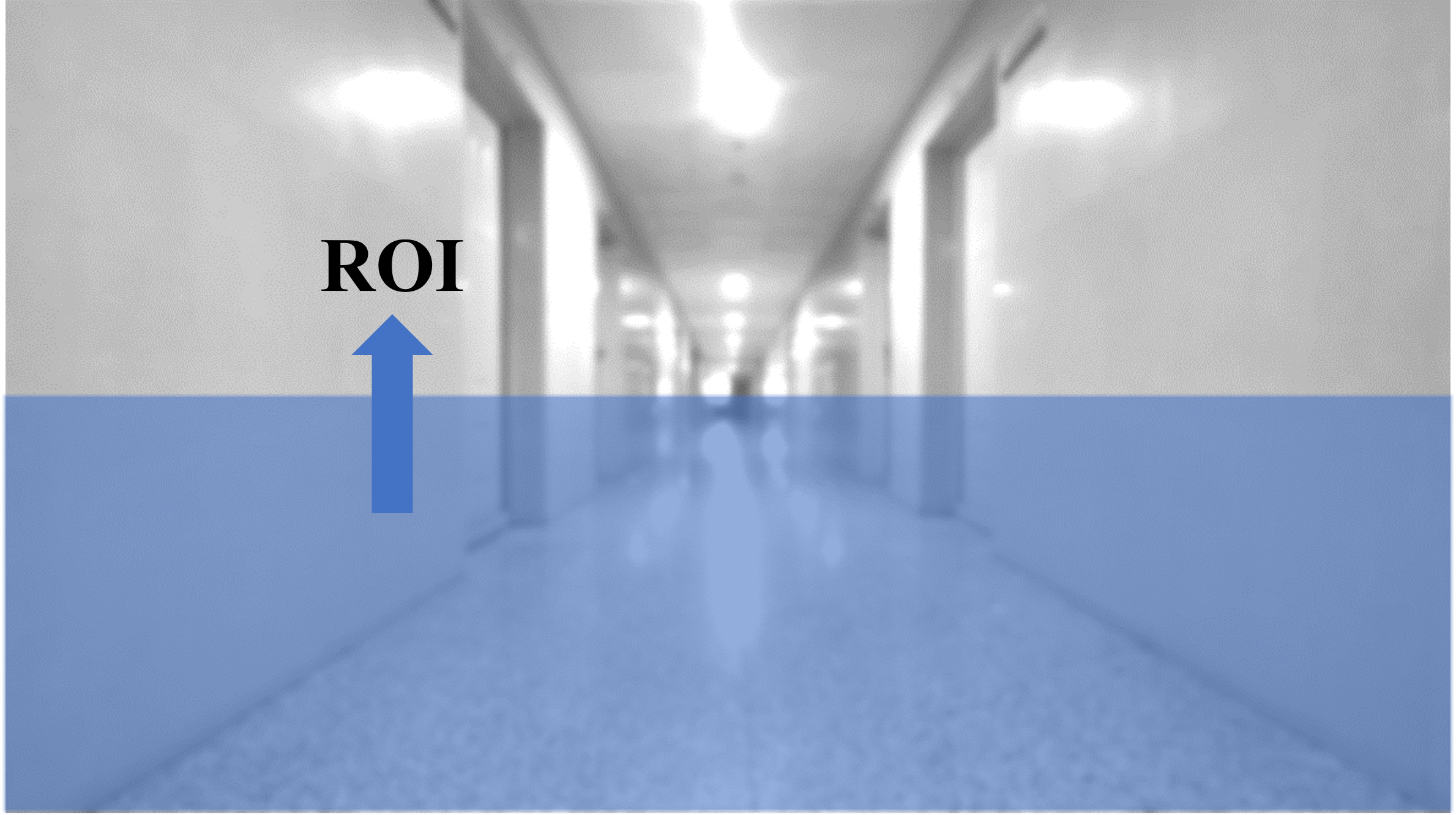}
	}
	\subfigure[\label{fig:b}]{
		\includegraphics[width=3cm]{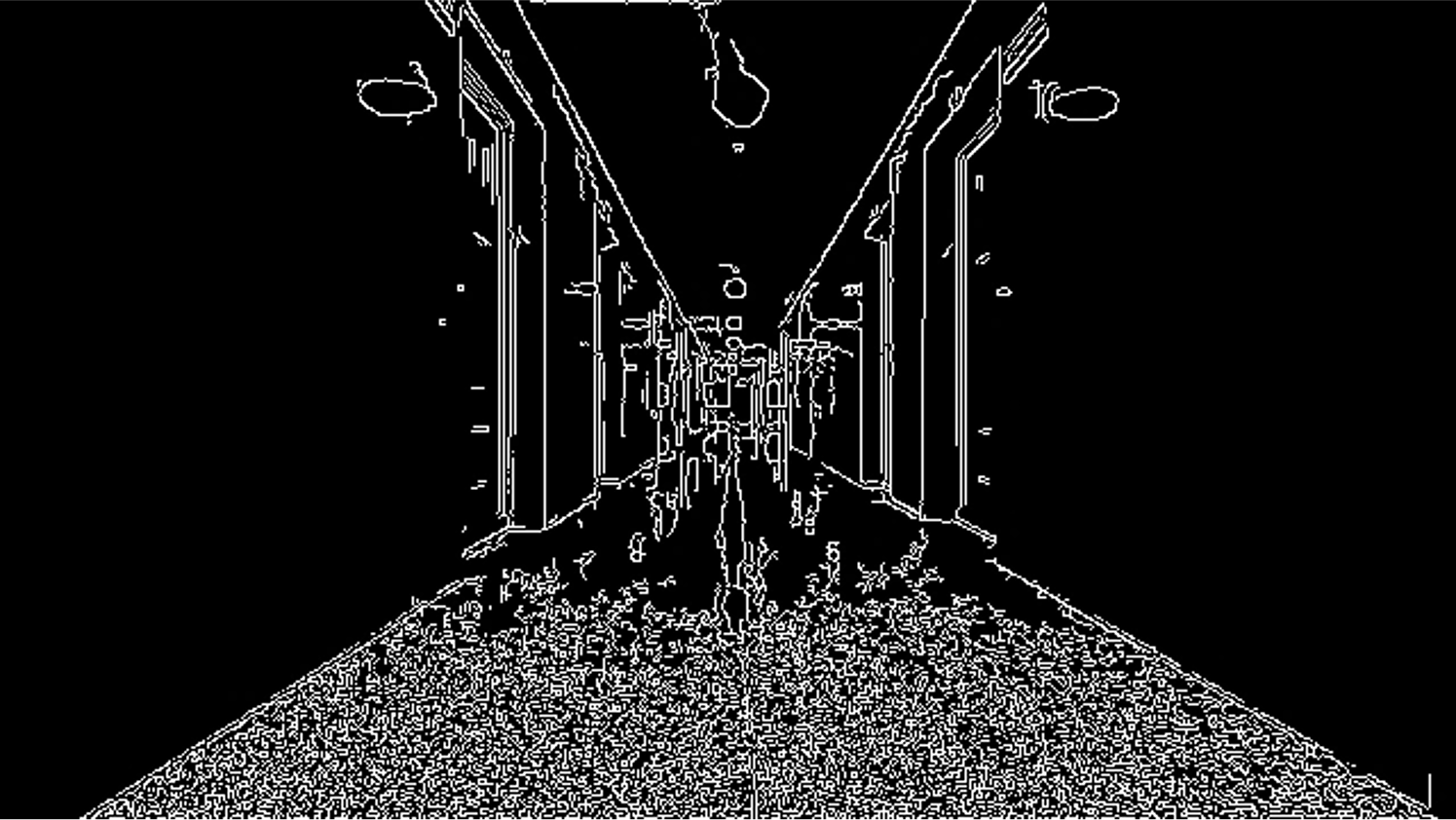}
	}
	\\
	\subfigure[\label{fig:c}]{
		\includegraphics[width=3cm]{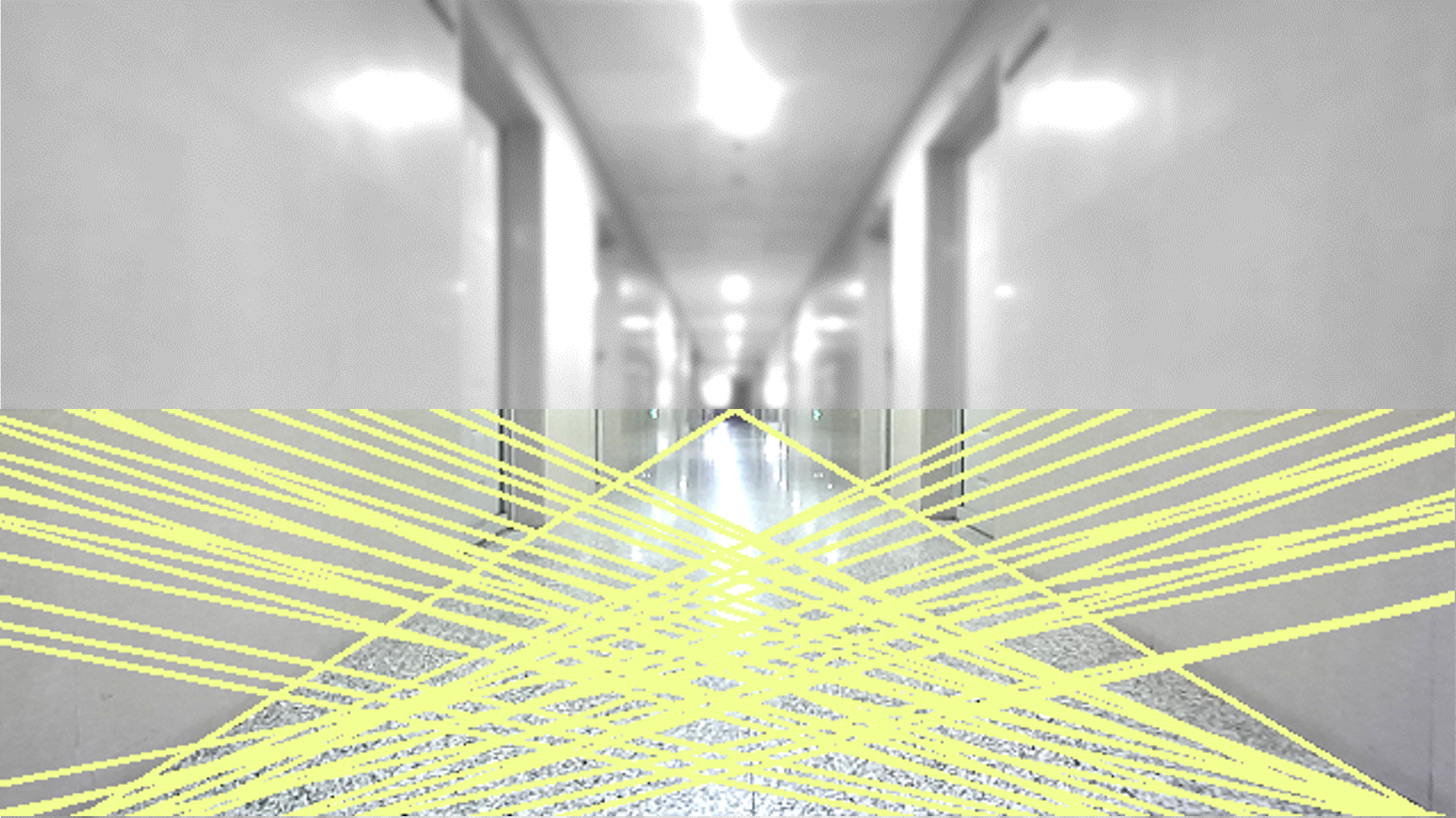}
	}
	\subfigure[\label{fig:d}]{
		\includegraphics[width=3cm]{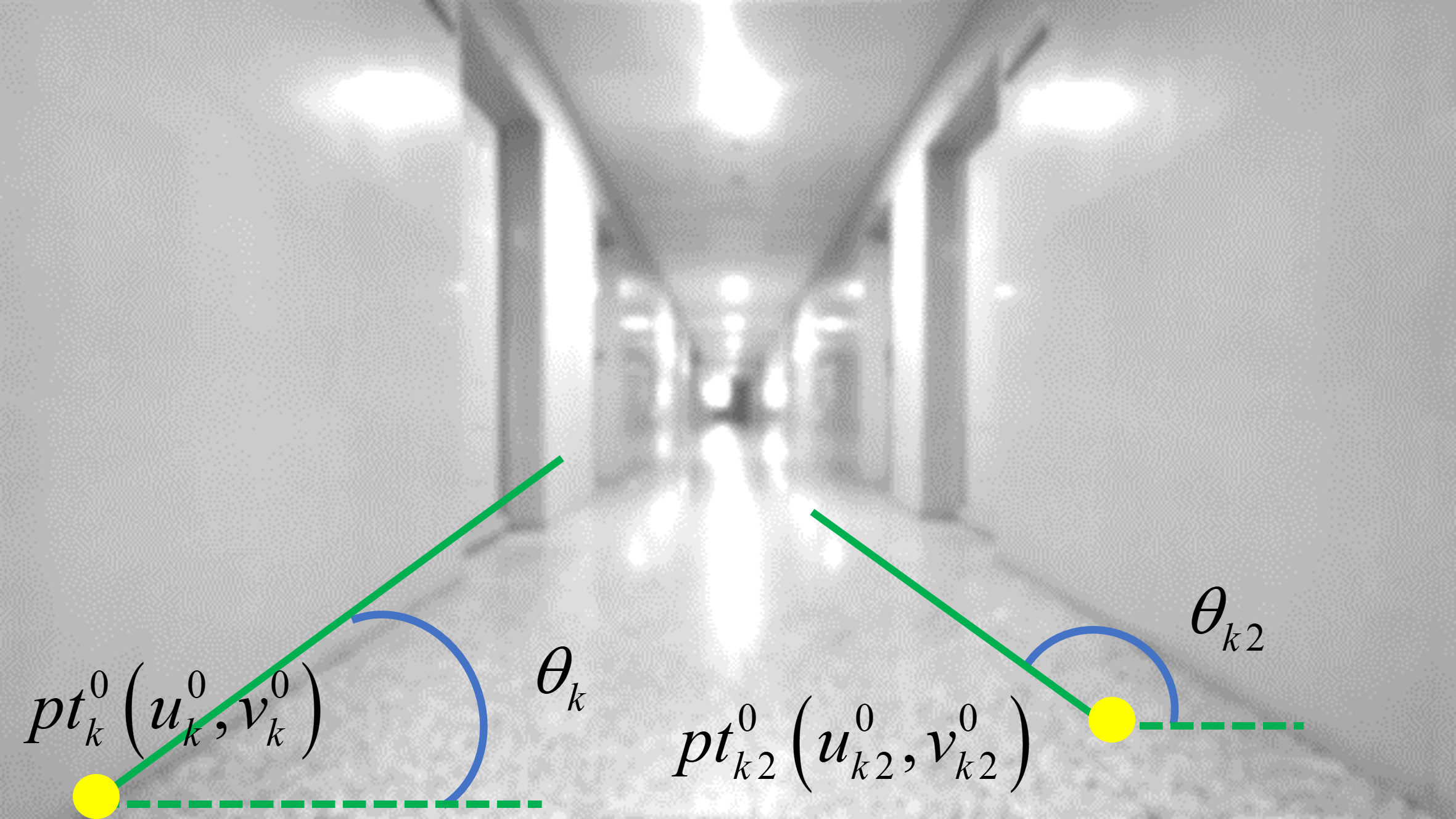}
	}
	\caption{(\textbf{a})The region of interestiong. (\textbf{b}) The results of Canny edge detection. 
		(\textbf{c}) The results of Hough line transform(after NMS). (\textbf{d}) The line definetion in the sence.\label{fig2}}
\end{figure}

\subsubsection{Construction of ROI}
In the context of long corridor/hallway scenarios, aside from ground edges, there often exist additional linear features such as door frames and objects, which can introduce interference with the detection of ground edge lines. Consequently, it becomes imperative to establish a Region of Interest (ROI) within the scene. This selection process is based on empirical observations. Due to geometric perspective, ground edge lines typically manifest in the lower-to-middle section of the image, converging from the bottom sides towards the center. Relying on this a priori knowledge, an ROI with a height of H within the image is designated, encompassing rows from the middle to the bottom of the image, as shown in Figure$~\ref{fig2}(\textbf{a})$.

\subsubsection{Hough Transform Based Line Feature Extration }
For the extraction process, this study employs a line feature detection algorithm based on the Hough transform. The Hough transform is a methodology designed to extract linear features from images. It leverages the duality between points and lines, mapping the discrete pixel points along a straight line in image space to curves in Hough space through parameter equations. Subsequently, the intersection points of multiple curves in Hough space are mapped back to straight line equations in image space, thereby forming the detected straight lines.

Before performing the Hough transform, the image is initially binarized and subjected to Canny edge detection 
thus expediting the process of Hough transformationas shown in Figure$~\ref{fig2}(\textbf{b})$. Subsequently, linear features are extracted through the Hough transform,as shown in Figure$~\ref{fig2}(\textbf{c})$, and the extracted features undergo linear fitting to yield the set of linear features, denoted as $S_l$.

	\begin{equation}
		S_l = \{l_k \left( {\theta}_k,{pt}_k^0 \right),k =1,2,...,n\}
	\end{equation}

Where,${\theta}_k,{pt}_k^0$are the inclination angle of $l_k$,and the the pixel coordinates of the starting point of $l_k$, respectively, as illustrated in Figure$~\ref{fig2}(\textbf{d})$.

\subsubsection{Scene-Prior Based Line Feature Extraction}
Based on scene priors, a reasonable range for setting the length and inclination angle of the lines is established, which is used to filter the lines within set $S_l$. If the image is strict symmetry of axis, the length of the edge lines $L\left( l_k \right)$ and the angle ${\theta}_k$ should satisfy the following conditions:

	\begin{equation}
		\begin{split}
			\frac{H}{4} \leq L\left(l_k\right) &\leq \frac{1}{2} \sqrt{H^2+W^2} 
			\\
			\arctan{\frac{H}{W}} \leq {\theta}_k < \frac{\pi}{2}  \;&\vee\; \frac{\pi}{2} < {\theta}_k \leq \arctan{-\frac{H}{W}}
		\end{split}
	\end{equation}

According to (2), the left and right line are picked out. And the edge line set is $S_{le} = \{l_l \left( {\theta}_l,{pt}_l^{0} \right) , l_r \left( {\theta}_r,{pt}_r^{0} \right)\}$, where $l_l$ and $l_r$ are left and right line, respectively.

\subsection{Visual Camera}
\subsubsection{Visual Camera Model}

The primary function of the virtual camera is to ensure consistency in the imaging process across different scenes, currently predominantly employed within deep learning-based computer vision methodologies. In 2023, BEV-LaneDet \cite{wang2022bev} introduced the concept of the virtual camera in the context of 3D lane detection tasks on the Bird's Eye View (BEV) plane. Due to variations in camera intrinsic parameters, installation positions, and camera poses, images captured by the same scene may have different dimensions and scaling ratios. The BEV-LaneDet method employs a deep neural network that requires image parameters to be as consistent as possible with those in the training dataset. This necessitates aligning the camera position, pose, and height above the ground during imaging to avoid substantial disparities between the predictive accuracy of the deployed model and that observed during offline training.

The virtual camera is manually defined, with its intrinsic parameters, installation position, and orientation preconfigured. Prior to training and inference with deep neural networks, images are initially projected onto the imaging space of the virtual camera through perspective transformation. This process ensures that the input images to the model exhibit consistency.

\begin{figure}
	\centering
	\includegraphics[width=8cm]{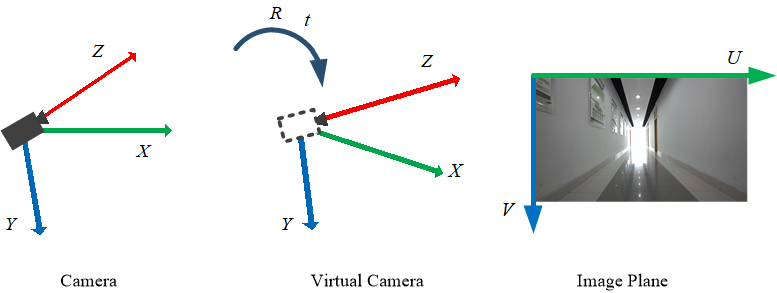}
	\caption{The visual camera defination and cordinate difination.\label{fig3}}
\end{figure}

In this study, the concept of the virtual camera is introduced to achieve algorithmic consistency across various scenes. The intrinsic coordinate system for both the real and virtual cameras is established as right-down-forward. The coordinate definitions for the two types of cameras and their respective images are depicted in Figure$~\ref{fig3}$. Within this hypothesis, the width of the long corridor remains constant, and the virtual camera is positioned at the center of the scene, equidistant from the left and right walls. Its optical axis is directed straight ahead along the length of the corridor, while maintaining consistent pitch angles, camera intrinsic parameters, and mounting height as the current real camera. 

To achieve the pose transformation from the real camera to the virtual camera, this study begins by projecting ground edge feature points onto the imaging plane of the virtual camera. Subsequently, a geometric error model is constructed, and the maximum likelihood estimation of the pose transformation is computed through an iterative optimization process.

\subsubsection{Virtual Camera Pose Estimation}
In the assumption of this study, the virtual camera is positioned at the center of the scene, ensuring that the edge lines in the image maintain its symmetry in the virtual camera. To fully utilize this structural feature and ensure algorithm consistency across different input images, the original image is initially transformed into the imaging space of the virtual camera. After obtaining the reference depth plane of the virtual camera, the depth plane information is then reprojected onto the current image. This process ultimately yields the depth distribution of the current image. Therefore, prior to conducting depth estimation, it is necessary to acquire the pose transformation from the current camera to the virtual camera.

Let $P^W$ represent a spatial point within the current scene, the pixel coordinate $P^C \left(u,v \right)$ of space point $P^W \left( x, y, z \right)$ is calculated as
	\begin{equation}
		\left[ \begin{array}{c}
			P^C \\ 1
		\end{array}\right] 
		= \left[ \begin{array}{c}
			u \\ v\\ 1
		\end{array}\right]
		=\frac{1}{z} \left[ \begin{array}{ccc}
			f_x & 0   & c_x \\
			0   & f_y & c_y \\
			0   &  0  &  1
		\end{array}\right]
		\left[ \begin{array}{c}
			x \\ y \\ z
		\end{array}\right]
		=\frac{1}{z} KP^W
	\end{equation}

Where, K is the camera internal parameter matrix, $f_x$ and $f_y$ are the pixel focal lengths in the x and y directions, respectively, corresponding to the physical focal length f of the camera. And the cx and cy represent the pixel offsets of the  optical center in the image of the camera. In general, the horizontal and vertical dimensions of the camera sensor have equal pixel sizes, in which case $f_x$=$f_y$. The rotation matrix R and displacement t are used to represent the pose transformation from the current camera to the virtual camera, and the current frame is projected onto the plane of the virtual camera. The projection result is calculated according to (3)

	\begin{equation}
		s^V P^C = K\left( RP^W + t \right)
	\end{equation}

Where $s^V$ is the scale in the transformed virtual camera. In the assumption of the virtual camera, the disparity in pose between the virtual camera and the real camera arises from the yaw angle $\phi$ and the displacement $\tau$ in the x-direction. In which case, R and t are calculated as 

	\begin{equation}
		R=
		\left[ \begin{array}{ccc}
			\cos{\phi}  &  0   & \sin{\phi}  \\
			0   &  1  & 0 \\
			-\sin{\phi}   &  0  &  \cos{\phi} 
		\end{array}\right]
		,
		t=
		\left[ \begin{array}{c}
			\tau \\ 0 \\ 0
		\end{array}\right]
	\end{equation}

As depth has not been recovered yet, the images lack scale information. In order to unify the scales in both cameras, a assumption is accepted of that for every pixel in the original image, its corresponding original spatial point lies in front of the camera at a distance of 1 meter. It means, for any spatial point $P^W \left( x, y, z \right)$, $z \equiv 1$. Therefore, the scale factor $s^V$ for the virtual camera is calculated as:

	\begin{equation}
		s^V = \cos{\phi} - \frac{u-c_x}{f}sin{\phi}
	\end{equation}

Based on equations (2) to (5), the transformation from the current frame image pixel point $pt^C(u,v)$ to the virtual camera image pixel point $pt^V(u^V,v^V)$ can be obtained as follows:

	\begin{equation}
		\begin{split}
			\left[ \begin{array}{c}
				u^V \\ v^V
			\end{array}\right]
			&=\xi
			\left[ \begin{array}{cc}
				cos\phi & 0 \\
				0  & 1
			\end{array}\right]
			\left[ \begin{array}{c}
				u \\ v
			\end{array}\right] \\
		\\
			&+\left[ \begin{array}{c}
				f_t \left(\tau + \sin{\phi}\right) + \left( \xi - cos{\phi}\right) \\
				\left(\xi + 1\right)c_y
			\end{array}\right]
			\\
		\end{split}
	\end{equation}

Where,$\xi ={1}/\left({f_t\cos{\phi}-u\sin{\phi}-c_x\cos{\phi}}\right)$,$f_t=f_x=f_y$.Each pixel of $l_l\left({\theta}_l,pt_l^0\right)$,$l_r\left({\theta}_r,pt_r^0\right)$along the edge can be transformed into the virtual camera according (7).

\begin{figure}
	\centering
	\includegraphics[width=5 cm]{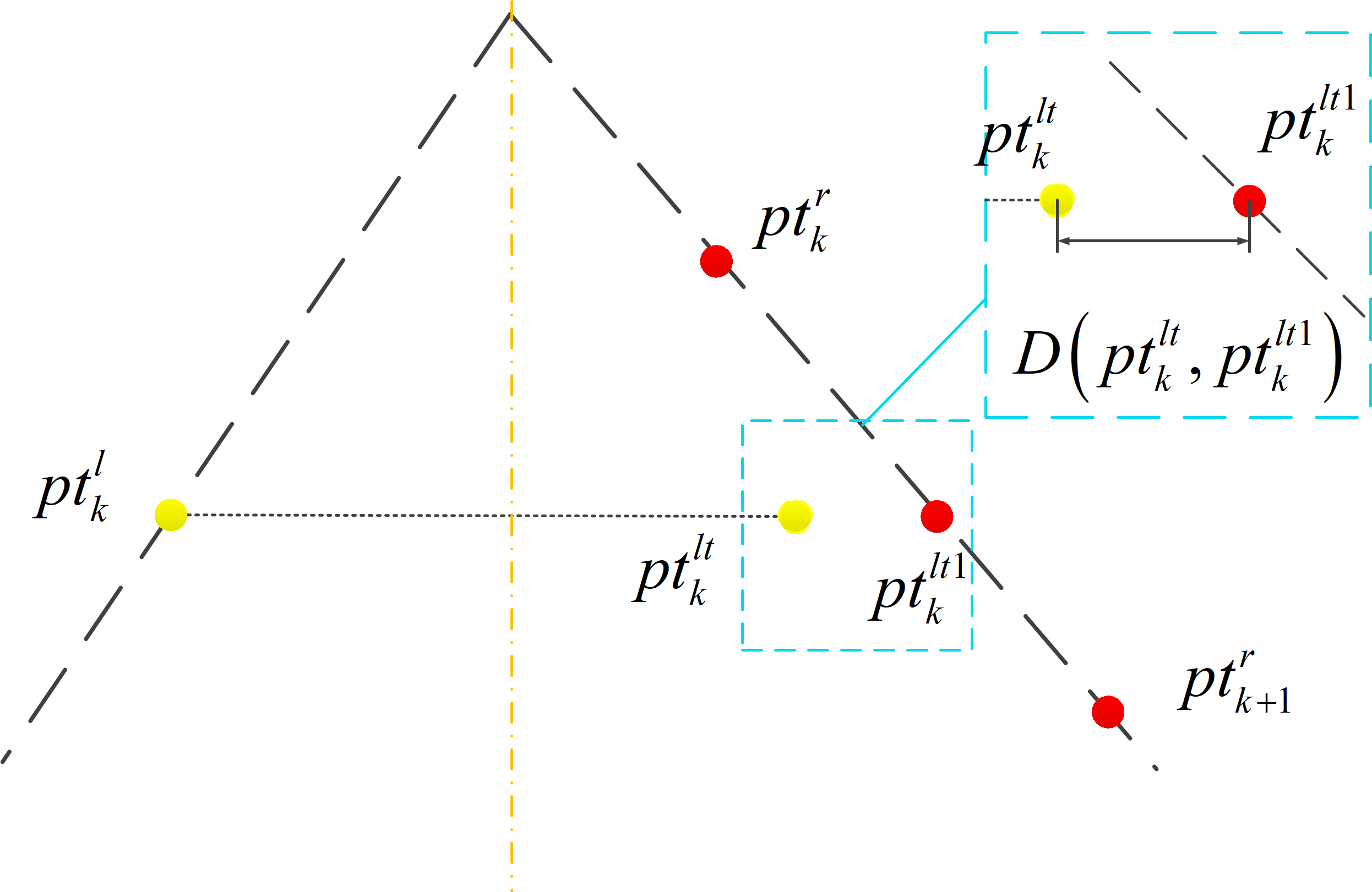}
	\caption{The symmetry geometric error.\label{fig4}}
\end{figure}

Within the assumption, the optical center of the virtual camera oincides with the symmetry axis of the corridor. As a result, the symmetry of the road edge lines is preserved in the virtual camera imaging space. The axis of symmetry is located at position $u=\frac{W}{2}$ on the image plane of virtual camera, where W represents the image width. Based on this structured feature, the geometric residual is calculated.
As shown in Figure$~\ref{fig4}$., Set $pt_k^{lt}=\left(u_k^l.v_k^l \right)$ as a point in $l_l$, and its symmetry point about $u=\frac{W}{2}$ is $pt_k^{lt}$.  Substituting $v = v_k^l$ into the equation of the right line, $pt_k^{lt1}$ can be obtained, and the distance D between two points can be calculated. According to (7), D is a function of both the yaw angle $\phi$ and the displacement $\tau$ in the x-direction. By uniformly selecting N feature points at intervals of $l_l$, the sum of D calculated for these N feature points yields symmetry error the left line $E_L$. Similarly, the  symmetry error of the right line $E_R$ can be obtained. Ultimately, this process yields the symmetry geometric error $E_G$.

	\begin{equation}
		E_G = E_L +E_R =\sum_{k}^{N}{D\left(pt_k^{lt},pt_k^{lt1} \right)} + \sum_{k}^{N}{D\left(pt_k^{rt},pt_k^{rt1} \right)}
	\end{equation}

Once the mathematical model for the symmetry-based geometric error $E_G$ is established, leveraging the prior features of the scene, a range of variability for the yaw angle $\phi$ and the displacement $\tau$ is defined. Nonlinear optimization is employed to compute the maximum likelihood estimation values $\widehat{\phi}$ and $\widehat{\tau}$ for the yaw angle $\phi$ and displacement $\tau$. Iterative calculations are performed within the range of variability for both parameters to minimize $E_G$. The resulting yaw angle and displacement that yield the smallest $E_G$ are considered as the maximum likelihood estimation results.

	\begin{equation}
		\widehat{\phi},\widehat{\tau} =argmin\left\{\sum_{k}^{N}{D\left(pt_k^{lt},pt_k^{lt1} \right)} + \sum_{k}^{N}{D\left(pt_k^{rt},pt_k^{rt1} \right)} \right\}
	\end{equation}

After obtaining the estimated value of the heading Angle and displacement of the heading Angle, the edge points are projected to the virtual camera space through the formula(4)(5), and $S^V=\left(pt_k^{Vl},pt_k^{Vr}\right),k=0,1,2... ,n$.

\subsection{Fast Monocular Dpth Recovery}
\subsubsection{3D Coordinates Estimation of Ground Edge Points}

In the virtual camera space, there are now pixel coordinates along two ground edges. Using the ground plane hypothesis, the points along each edge are assumed to be coplanar, and the 3D spatial coordinates of each point meet $\forall P^W_k\left( x_k,y_k,z_k \right) \in S^W, y_k = 0$. According to the installation position and aperture Angle information of the camera, the 3D spatial coordinates of each edge point corresponding to the virtual camera can be solved by geometric method. The depth plane can then be constructed.

\begin{figure}
	\centering
	\includegraphics[width=8 cm]{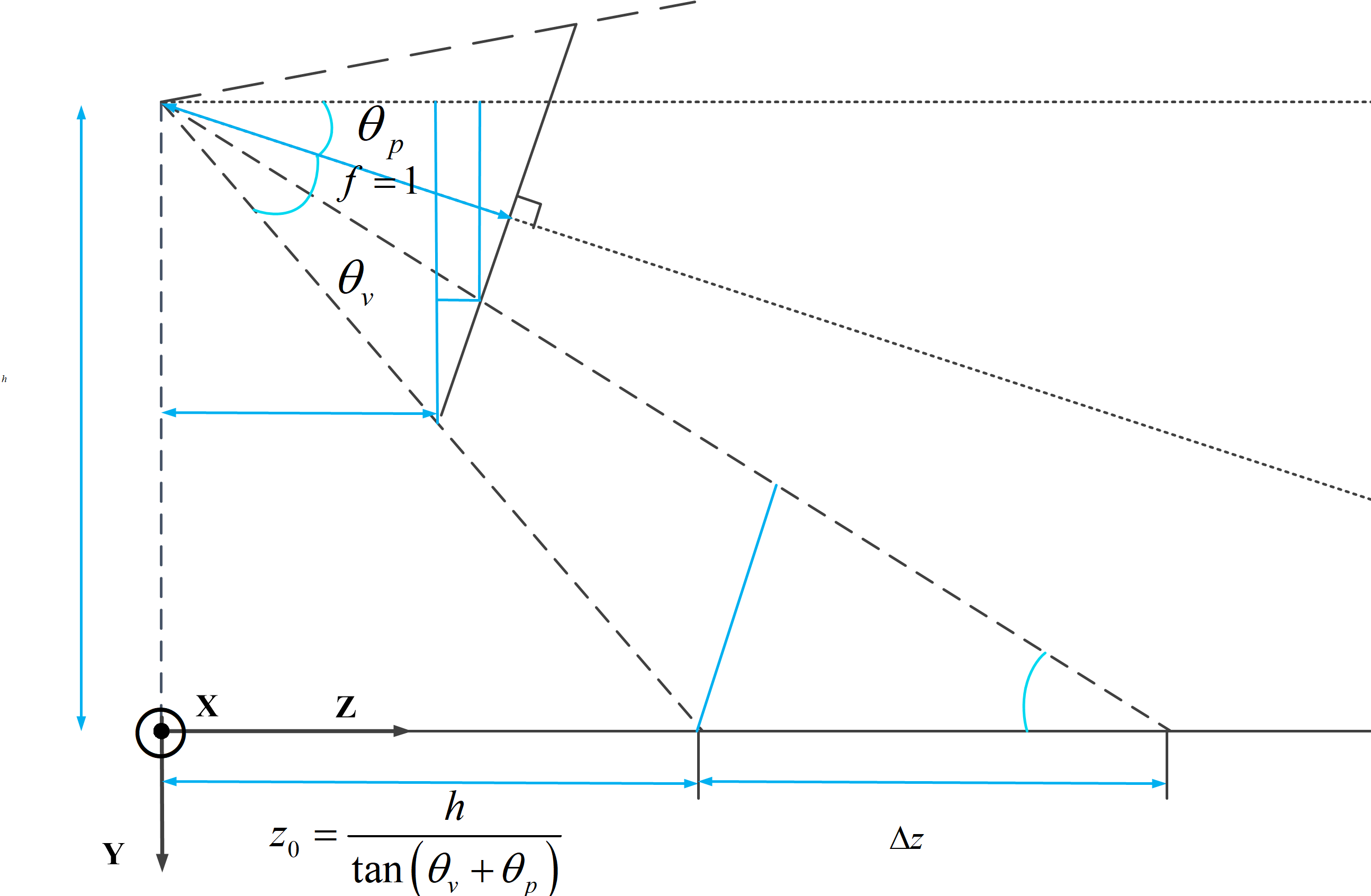}
	\caption{The geometry of the $\Delta z$ computation.\label{fig5}}
\end{figure}

The space coordinate system is selected as the lower left front, the camera height is known as h, the vertical aperture Angle is $\theta_v$ . Assume that the pitch Angle of the virtual camera in the current scene is $\theta_p$ , and for a pixel point  $pt_0$ at the bottom of the image, its pixel coordinate is $\left( u_t,v_t\right)$   meet $0 \leq u_t \leq W$,$v_t = H$ , W,H are the image height and image width respectively, and its depth is calculated as

	\begin{equation}
		z_0 = \frac{h}{\tan{\left(\theta_v + \theta_p\right)}}
	\end{equation}

The theoretical method of Inverse Perspective Transform (IPM) is used to calculate the depth of each edge point, as shown in Figure $~\ref{fig5}$. For any point in the set of edge points $S^V=\left(pt_k^{Vl},pt_k^{Vr}\right),k=0,1,2... ,n$, its pixel coordinate is $\left( u_k^V,v_k^V\right)$, and its depth is calculated according to the inverse perspective transform.

	\begin{equation}
		z =  z_0 + \Delta z
	\end{equation}

The equation is constructed to solve   according to the perspective geometry as (12)

	\begin{equation}
		\frac{\eta_k\cos{\left(\theta_k - \theta_p\right)}}{\Delta z \sin{ \theta_p}} = \frac{z_1}{z_0}
	\end{equation}

Where , $ \eta_k = f_y \Delta v_k , \Delta v_k = H - v_k^V$is the depth in space at the bottom of the imaging plane 1m away from the optical center of the camera, 

	\begin{equation}
		z_1 = cos\theta_v - sin\theta_v sin\theta_p
	\end{equation}

Solving (9),$ \Delta z $ is calculated as

	\begin{equation}
		\Delta z = z_0 \frac{z_0 \cos{\theta_p} + h\sin{\theta_p} }{ z_1 h - \eta_k z_0 \cos{ \theta_p}} \eta_k
	\end{equation}

\subsubsection{3D Coordinates Optimization $ \& $ Pitch Angle Estimation}

As shown in (14), the calculation of edge point depth $ \Delta z $ is related to camera elevation Angle $ \theta_p $. In this paper, $ \Delta z $ and $ \theta_p $ are estimated simultaneously by nonlinear optimization. In the corridor, the road surface in the scene is flat and the pitch Angle variation range is small. The iterative optimization algorithm is designed to iteratively calculate the geometric residual of 3D space points within the interval $ \left[ \lambda_1 , \lambda_2 \right] $, taking $ \alpha $ as the step length, and calculate the optimal $\theta_p$,to estimate the result.

Set the initial value of $\theta_p$, according to (14) for the edge point pixel set $ S^V $, in which all pixels recover the depth distribution, according to camera imaging model (3), the 3D spatial coordinates of edge point $ pt_k^{Vl}   $ are calculated as

	\begin{equation}
		\left\{
		\begin{aligned}
			&z^W_k = z_0 + \Delta z_k 
			\\
			&x_k^W = z_k \left( u_k^V - c_x \right)/f_t
			\\
			&y_k^W = z_k \left( v_k^V - c_y \right)/f_t
		\end{aligned}
		\right.
	\end{equation}

And the geometry residual is calculated as
	\begin{equation}
		\begin{aligned}
			E_G^W &= \sum_{k}^{L}{\left\| P_k^{Wl} , P_k^{Wr} \right\| _2}
			\\
			&\Longleftrightarrow \sum_{k}^{L}{\left| x_k^{Wl} , x_k^{Wr} \right|}_{y_k^{Wl} = y_k^{Wr}}
			\\
			&=\sum_{k}^{L}{\frac{z_k}{f_t} \left(1+\frac{z_0 \cos{\theta_p} + h\sin{\theta_p} }{ z_1 h - \eta_k z_0 \cos{ \theta_p}} \eta_k \right)}\left(u_k^{Vl}-u_k^{Vr}\right)
		\end{aligned}
	\end{equation}

The approximate optimal estimate of $\theta_p$ can be obtained by iterative optimization

	\begin{equation}
		\hat{\theta}_p = \arg\min
		\left\{ \sum_{k}^{L}{\frac{z_k}{f_t} \left(1+\frac{z_0 \cos{\theta_p} + h\sin{\theta_p} }{ z_1 h - \eta_k z_0 \cos{ \theta_p}} \eta_k \right)}\left(u_k^{Vl}-u_k^{Vr}\right)\right\}
	\end{equation}

\subsubsection{Depth Plane Construction $\&$ Spatial Point Depth Recovery}
In the image space of the virtual camera, the spatial points in the same depth plane retain the parallel characteristics, and the contour points of the same depth are connected to build a depth plane. The depth plane $ \Gamma_k^V $ is guided by the contour lines. The depth plane $ \Gamma_k^V $ is determined by the left and right edge points $ \left\{pt_k^{Vl}\left( u_k^{Vl},v_k^{Vl}\right),pt_k^{Vr}\left( u_k^{Vr},v_k^{Vr}\right) \right\}$, and is uniquely determined, $Dp \left(\Gamma_k^V \right) $ is the depth of  $ \Gamma_k^V $, $Dp \left(\Gamma_k^V \right) = z_k$.

The depth plane $ \Gamma_k^V $, obtained in the virtual camera imaging space, is inversely transformed according to (4-5), and the set of depth planes in the real camera image can be obtained: $ S_{\Gamma}^C = \left\{ \Gamma_k^C \right\} ,k=1,2,3,...,L$. In a real camera image, if the pixel $ pt_k^{C}\left( u_k^{C},v_k^{C}\right) $ belongs to the depth plane  $  \Gamma_k^C $ then equation (18) is a necessary condition.

	\begin{equation}
		\forall pt_k^{C} \in \Gamma_k^C \Rightarrow 
		\left\{
		\begin{aligned}
			& f\left(u_k^{C}\right) \leq u_i^C \leq u_k^{C} 
			\\
			& v_k^{Cl} \leq v_i^C \leq v_k^{Cr}
			\\
			& u_{k-1}^{C} \leq u_i^C  \;\vee\; v_i^C \leq v_{k-1}^{Cl}  \;\vee\; v_i^C \geq v_{k-1}^{Cr}
		\end{aligned}
		\right.
	\end{equation}

Where $ f\left(u_k^{Cl}\right) = \alpha u_k^{Cl} -\beta $ is a linear transform of $ u_k^{Cl} $ , and $ \alpha $ , $ \beta $ are empirical coefficients.

In the corridor scene, the spatial point distribution is relatively ideal, ignoring the influence of dynamic objects, occlusion, etc., and taking (18) as sufficient and necessary conditions of $pt_k^{cl} \in \Gamma_k^C$ , each pixel in the image is classified.

After pixel classification is completed, it is determined whether it is a ground point according to the coordinates of each pixel. The condition that $ pt_i^{V }\left( u_i^{V },v_i^{V }\right) $ is a ground point is

	\begin{equation}
		pt_i^{C} \in \Gamma_k^C   \;\wedge\; u_{k-1}^{C} \leq u_i^C  \leq u_{k-1}^{C}
	\end{equation}

For the pixel point $ pt_k^{C}\left( u_k^{C},v_k^{C}\right) \in  \Gamma_k^C $, its depth is calculated by interpolation method (20).

	\begin{equation}
		Dp\left( pt_i^{C} \right) =Dp \left(\Gamma_k^C \right) +\sigma \left( Dp \left(\Gamma_{k-1}^C\right)  - Dp \left(\Gamma_k^C\right) \right)
	\end{equation}

In the formula, $ \sigma $ is a nonlinear coefficient. When the depth plane is sufficiently dense, linear interpolation method is adopted. If the point is a ground point, then

	\begin{equation}
		\sigma = \frac{u_{i}^{C}-u_{k-1}^{C}}{u_{k }^{C}-u_{k-1}^{C}}   
	\end{equation}

Otherwise,

	\begin{equation}
		\sigma = \left\{
		\begin{aligned}
			&\frac{v_{i}^{C}-v_{k-1}^{Cl}}{v_{k }^{Cl}-v_{k-1}^{Cl}} \, , \, v_{i}^{C} \leq v_{k-1}^{Cl}
			\\
			& \frac{v_{i}^{C}-v_{k-1}^{Cl}}{v_{k }^{Cr}-v_{k-1}^{Cr}} \, , \, v_{i}^{C} \geq v_{k-1}^{Cr}
		\end{aligned}
		\right.
	\end{equation}

\section{Results}

\subsection{Experiment Overview}

\begin{figure}
	\centering
	\includegraphics[width=8cm]{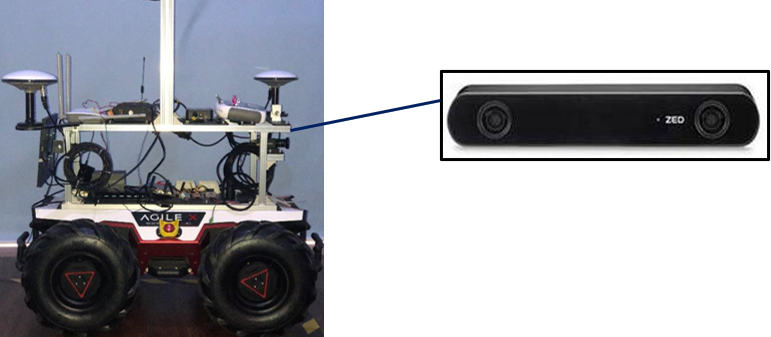}
	\caption{The Unmanned Ground Vehicle(UGV) and ZED2 camera}
\end{figure}
\begin{figure}
	\centering
	\subfigure[\label{fig:a}]{
		\includegraphics[width=3cm]{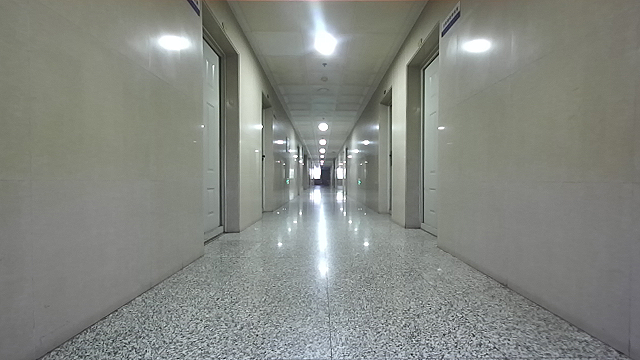}
	}
	\subfigure[\label{fig:b}]{
		\includegraphics[width=3cm]{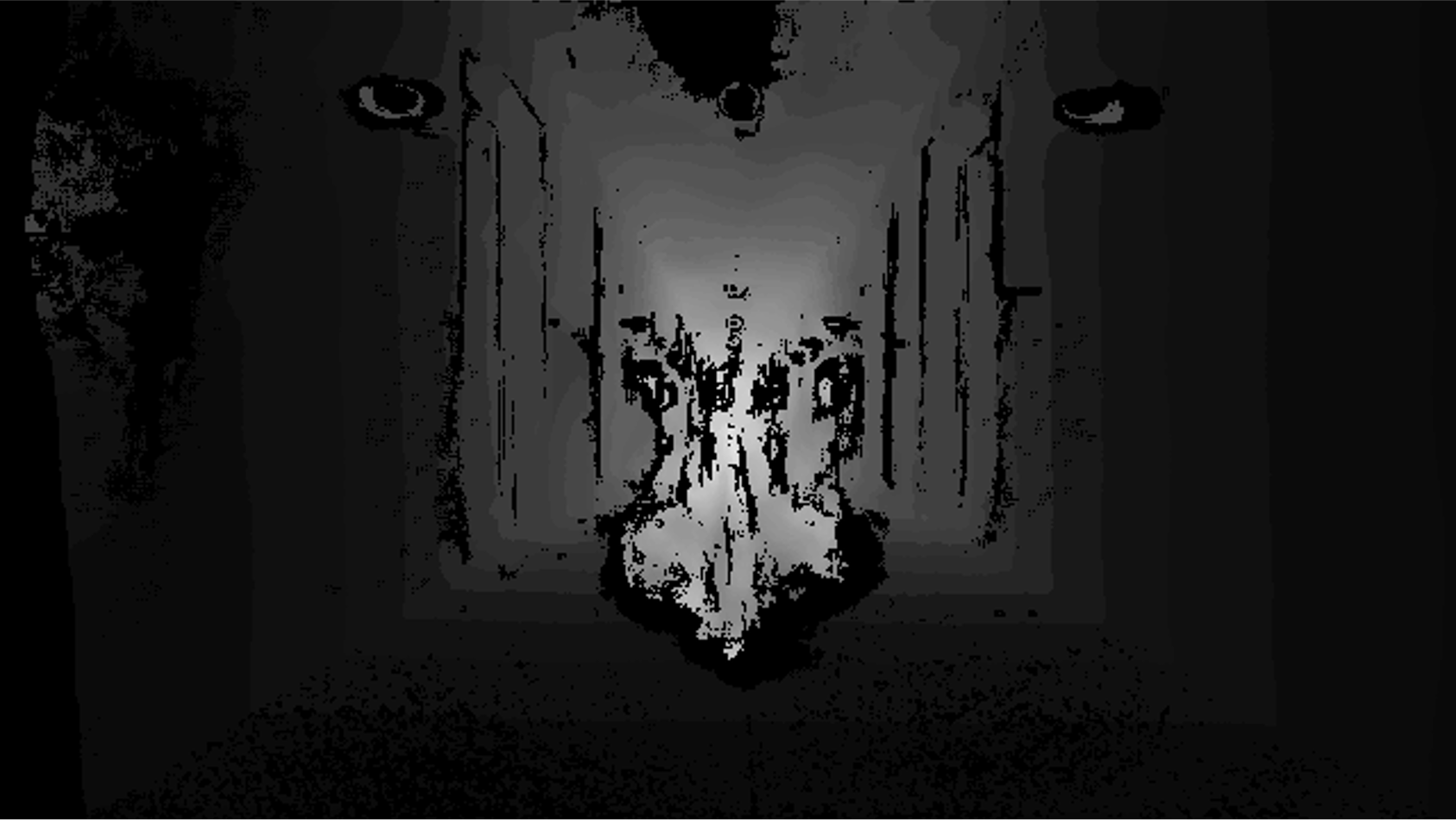}
	}
	\\
	\subfigure[\label{fig:c}]{
		\includegraphics[width=3cm]{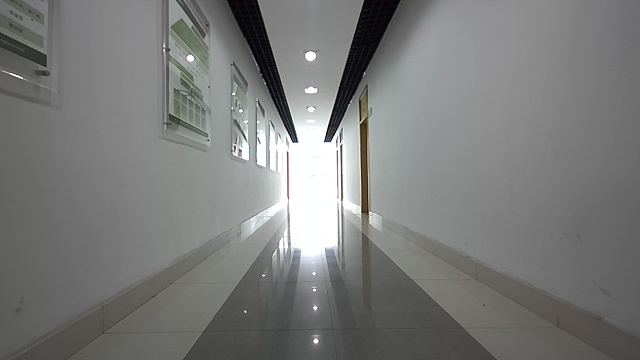}
	}
	\subfigure[\label{fig:d}]{
		\includegraphics[width=3cm]{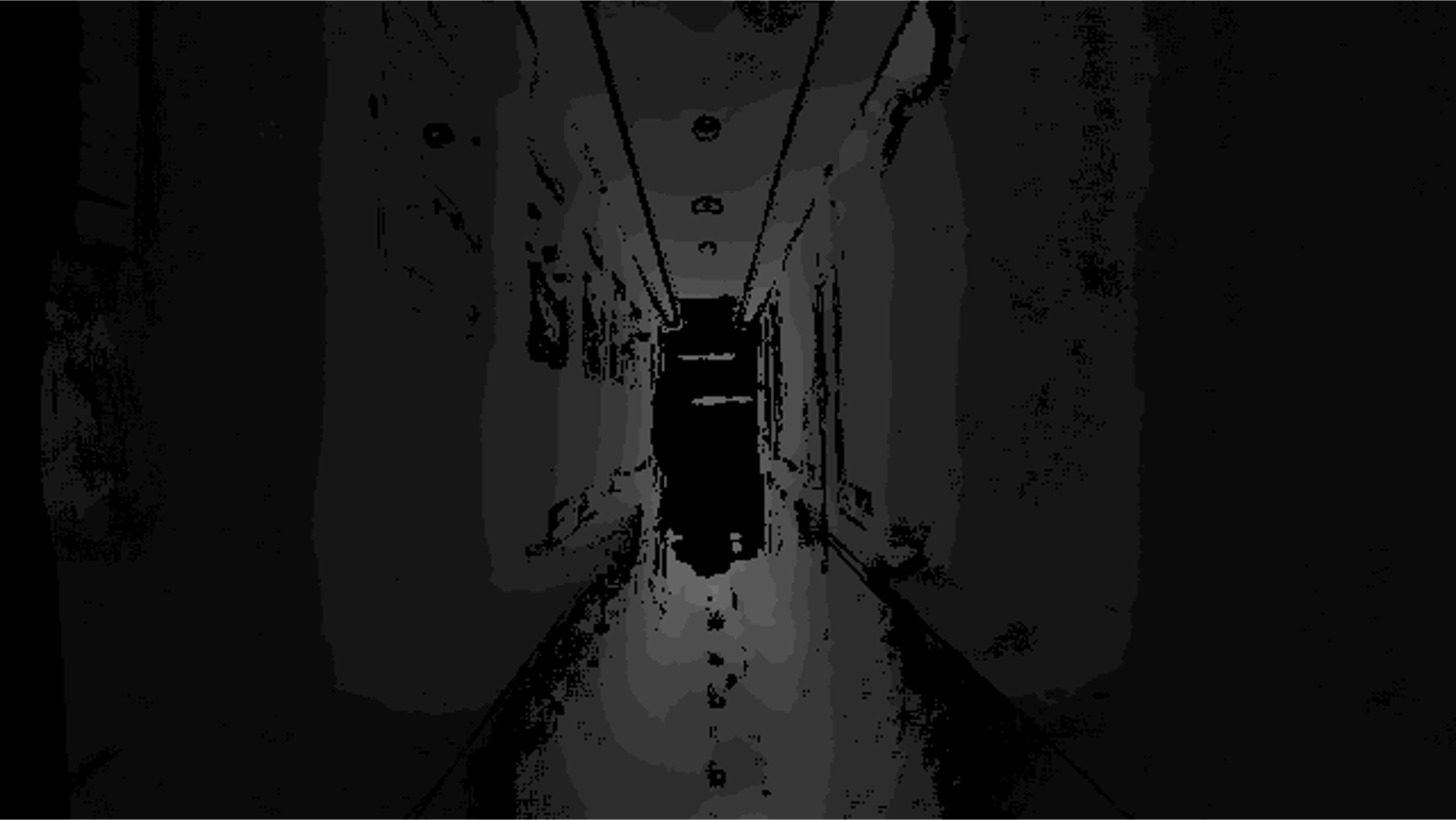}
	}
	\caption{(\textbf{a})Source img of Cord\_E05\_z sence. (\textbf{b}) Depth truth of Cord\_E05\_z sence. 
		(\textbf{c}) Source img of Cord\_H02\_z sence. (\textbf{d}) Depth truth of Cord\_H02\_z sence.\label{fig6}}
\end{figure}

  
The algorithm proposed in this paper is validated by collecting data in real scenarios. The ZED2 camera sensors mounted on an Unmanned Ground Vehicle(UGV) were used to collect images in different scenes to build data sets, and the RGB images and 16-bit depth images output by the cameras were recorded by ROSbag. The size of the output RGB image and depth image is 420x360 for the ZED2 camera. Camera mounting height are 0.66m, 0.62m respectively. The experimental equipment is shown Figure $~\ref{figt}$
ZED2 camera is used to collect 135 images of 9 kinds corridor with length range $[0-100]m$ and width range $[2-4]m$ and different lighting conditions, from which a corridor dataset, named Corr\_EH\_z were constructed. There are two parts of Corr\_EH\_z, $Cord\_Exx\_z$ , $Cord\_Hxx\_z$. The $Cord\_Exx\_z$  subset is a simple condition while $Cord\_Hxx\_z$ is a complex condition.  In Figure $~\ref{fig6}$, scene images and corresponding depth images in two types of subsets are shown respectively.

\subsection{Experiment 1: Verification of Algorithm Accuracy}
In the two types of sub-datasets, the heading Angle estimation interval is set as $[-0.314,0.314]rad(±18°)$ interval, and the accuracy of the pitch Angle estimation and corridor width estimation in this paper is tested with 0.05rad as the step length. The estimation results are shown in Table 1. The relative error of the output corridor width estimation value of the method in this paper is less than 0.0427. The mean relative error is 0.0221.

\begin{table}[h]
	\caption{Accuracy of corridor width estimation.\label{tab1}}
	\label{table_example}
	\begin{center}
	\begin{tabular}{cccc}
		\hline
		\textbf{Scene}	& \textbf{Estimated(m)}	& \textbf{Groundtruth(m)} & \textbf{relative error \%}\\
		\hline
		Cord\_E01\_z		& 2.16			& 2.11 &  2.3697\\
		Cord\_E02\_z		& 2.08			& 2.13 &  2.3474\\
		Cord\_E03\_z		& 2.20			& 2.11 &  4.2654\\
		Cord\_E04\_z		& 2.17			& 2.09 &  3.8278\\
		Cord\_E05\_z		& 1.98			& 2.02 &  1.9802\\
		Cord\_E06\_z		& 1.88			& 1.86 &  1.0753\\
		Cord\_H01\_z		& 2.16			& 2.13 &  1.4085\\
		Cord\_H02\_z		& 3.04			& 3.09 &  1.6181\\
		Cord\_H03\_z		& 2.95			& 2.98 &  1.0018\\
		\hline
	\end{tabular}
	\end{center}
	\noindent{\footnotesize{\textsuperscript{1} Tables may have a footer.}}
\end{table}

The depth estimation accuracy of the proposed method is tested in the $Cord\_Exx\_z$ and $Cord\_Hxx\_z$ datasets, as shown in Table $~ref{tab2,tab3}$. Four types of general accuracy indexes of the proposed method are calculated under the two evaluation conditions of depth truth value and depth truth value respectively, and the four types of accuracy indexes are defined as \cite{eigen2014depth}

	\begin{equation*}
		\begin{aligned}
			&AbsRel: &\frac{1}{N}\sum_{p}^{N}\left|\frac{y_p-{\hat{y}_p}}{y_p}\right|
			\\
			&Log10: &\frac{1}{N}\left|\log_{10} {y_p}-\log_{10} {\hat{y}_p}\right|
			\\
			&RMSE: &\sqrt{\frac{1}{N}\sum_{p}^{N}\left({y_p}- {\hat{y}_p}\right)^2}
			\\
			&RMSElog: &\sqrt{\frac{1}{N}\sum_{p}^{N}\left(\log_{10}{y_p}-\log_{10} {\hat{y}_p}\right)^2}
		\end{aligned}
	\end{equation*}

\begin{table}[h]
	\caption{Accuracy of depth estimation while depth is less than 5m.\label{tab2}}
	\begin{center}
		\begin{tabular}{ccccc}
			\hline
		\textbf{Scene}	& \textbf{AbsRel}	& \textbf{Log10} & \textbf{RMSE} & \textbf{RMSElog}\\
			\hline
			Cord\_E01\_z		& 0.06338 &0.02908 &0.30892 &0.05119\\
			Cord\_E02\_z		& 0.06094 &0.02630 &0.26311 &0.04062\\
			Cord\_E03\_z		& 0.05813	&0.02522	&0.27610	&0.04183 \\
			Cord\_E04\_z		& 0.07106	&0.03279	&0.32257	&0.05431\\
			Cord\_E05\_z		& 0.06011	&0.02518	&0.35607	&0.04817 \\
			Cord\_E06\_z		& 0.07028	&0.03156	&0.32098	&0.05422\\
			Cord\_H01\_z		& 0.08588	&0.04104	&0.45468	&0.07382\\
			Cord\_H02\_z		& 0.11233	&0.05968	&0.65776	&0.11828\\
			Cord\_H03\_z		& 0.10867	&0.04709	&0.46969	&0.06848\\
			\hline
		\end{tabular}
	\end{center}
	\noindent{\footnotesize{\textsuperscript{1} Tables may have a footer.}}
\end{table}

\begin{table}[h]
	\caption{Accuracy of depth estimation while depth is less than 40m.\label{tab3}}
	\begin{center}
		\begin{tabular}{ccccc}
			\hline
		\textbf{Scene}	& \textbf{AbsRel}	& \textbf{Log10} & \textbf{RMSE} & \textbf{RMSElog}\\
		\hline
		Cord\_E01\_z& 0.09740	& 0.05185	& 1.26336	& 0.11411 \\
		Cord\_E02\_z& 0.08256	& 0.04079	& 1.04763	& 0.09274 \\
		Cord\_E03\_z& 0.09026	& 0.01348	& 1.29909	& 0.11610 \\
		Cord\_E04\_z& 0.10603	& 0.05551	& 1.89761	& 0.12333 \\
		Cord\_E05\_z& 0.08386	& 0.04342	& 1.28335	& 0.10847 \\
		Cord\_E06\_z& 0.09879	& 0.05225	& 1.25565	& 0.11414 \\
		Cord\_H01\_z& 0.12416	& 0.07247	& 1.61772	& 0.15947 \\
		Cord\_H02\_z& 0.16391	& 0.10121	& 1.57186	& 0.19957 \\
		Cord\_H03\_z& 0.12647	& 0.06797	& 1.43041	& 0.14808 \\
		\hline
	\end{tabular}
	\end{center}
	\noindent{\footnotesize{\textsuperscript{1} Tables may have a footer.}}
\end{table}

As can be seen from Table $~\ref{tab2}$, under the condition of depth truth value <5m, the AbsRel index $< 7.106\%$ and RMSE index $<0.35607$ output depth estimates in the $Cord\_Exx\_z$ dataset of this method reach the advanced level of accuracy. At the same time, the accuracy of the proposed method can be predicted in the depth range of 40m, as shown in Table $~\ref{tab3}$. In the $Cord\_E03\_z$ scenario, the AbsRel of the proposed method in the depth range of 40m can reach $8.256\%$. Figure $~\ref{fig7}$ shows the depth recovery effect of the proposed method in several scenario.It should be noted that the ceiling part is eliminated to save computing resources.

\subsection{Experiment 2: Comparison with the State-of-the-art Method}
In the $Cord\_Exx\_z$ dataset, a precision comparison test was conducted between the proposed method and the ADABINS\cite{bhat2021adabins} method based on deep learning. ADABINS method is based on codec-decoding network +ViT for depth classification estimation. This method was published in CVPR in 2021, and currently ranks 14th in Kitti\cite{geiger2013vision} list and 25th in NYU\cite{silberman2012indoor} list, which is at the advanced level among existing methods.

First, 25epochs of ADABINSS method was trained on $Cord\_Exx\_z$ sub-dataset, and then the accuracy of depth prediction between the proposed method and the ADABINS method was compared, as shown in Table$~\ref{tab4}$.

\begin{table}[h] 
	\begin{center}
	\caption{Precision comparison experiment with Adabins.\label{tab4}}
	\begin{tabular}{ccccc}
		\hline
		\textbf{Method}	& \textbf{AbsRel}	& \textbf{Log10} & \textbf{RMSE} & \textbf{RMSElog}\\
		\hline
		ADABINS\textsuperscript{1}& 0.079	& 0.033	& 0.299	& 0.101 \\
		Ours\textsuperscript{1}& 0.083	& 0.036	& 0.363	& 0.053 \\
		ADABINS\textsuperscript{2}& -	& -	& -	& - \\
		Ours\textsuperscript{2}& 0.098	& 0.054	& 1.425	& 0.121 \\
		\hline	
	
	\end{tabular}
	\end{center}
	\noindent{\footnotesize{\textsuperscript{1} Depth range 0-5m.} \\ \footnotesize{\textsuperscript{2} Depth range 0-40m.}}
\end{table}

AbsRel, Log10 and RMSE accuracy indexes of the proposed method are similar to those of ADABINS under the condition of depth truth value <5m, and RMSElog is 0.053, which is better than ADABINS (0.101). AbsRel and RMSE were $9.8\%$ and 1.425 respectively under the condition of depth truth value <40m, while the effective depth prediction range of ADABINS indoor depth estimation is only 10m.

Compare the speed of reasoning/recovering a single image between ADABINS and the proposed method in different computing platforms. Computing platform 1 is AMDThreadripperPRO 5975W 32-core high-performance processor with single-core main frequency 3.6GHz, and GPU is Nvidia RTX4090 graphics processing unit, which is used to assist deep neural network reasoning. Computing platform 2 is a medium-low performance Intel Core i5-7300HQCPU with a main frequency of 2.5GHz; Repeated experiments were used to calculate the average execution time of the two methods, and the results were shown in Table $~\ref{tab5}$.

\begin{table}[h] 
	\begin{center}
	\caption{Running time comparison experiment with Adabins.\label{tab5}}
	\begin{tabular}{cc}
		\hline
		\textbf{Method}	& \textbf{Average running time}	\\
		\hline
		ADABINS\textsuperscript{1}& 0.0507	\\
		Ours\textsuperscript{1}& 0.0097		\\
		ADABINS\textsuperscript{2}& -	\\
		Ours\textsuperscript{2}& 0.0482	 \\
		\hline
	\end{tabular}
	\end{center}
	\noindent{\footnotesize{\textsuperscript{1} On platform 1.} \\ \footnotesize{\textsuperscript{2} On platform 2.}}
\end{table}

The average execution time of the proposed method for depth recovery of a single image on computing platform 1 is 0.0097s, which is less than 1/5 of the average execution time of the ADABINS method under GPU acceleration, and the average execution time of the proposed method on processor 2 is 0.048s. The results show that the proposed method can achieve a real-time processing speed of 20FPS in low and medium performance processors.

\section{Discussion}
In this paper, an explicit method for fast monocular depth recovery in long corridor scenes is proposed. 

By extracting key information in long corridor scenes, the method optimizes pitch angle estimates and scene edge point depths by minimizing geometric residuals, and classification of space points by constructing depth planes, the monocular depth estimation problem is transformed into a solvable optimization problem. Fast monocular depth estimation for long corridor scenarios is achieved. In this paper, we collected the corridor image construction datasets under various scenarios, tested the performance of this method experimentally, and conducted precision comparison test with the method based on deep learning. The test results show that:

\begin{enumerate}
	\item	The accuracy of the explicit method for fast monocular depth recovery in long corridor scenarios reaches the advanced level of the existing monocular depth estimation algorithms;
	\item	The fast monocular depth recovery method greatly accelerates the depth recovery process of a single image. In low and medium performance processors, this method can perform real-time depth estimation of corridor scenes at a speed of 20FPS.
\end{enumerate}

\begin{figure}[h]
	\begin{center}

	\includegraphics[width=6 cm]{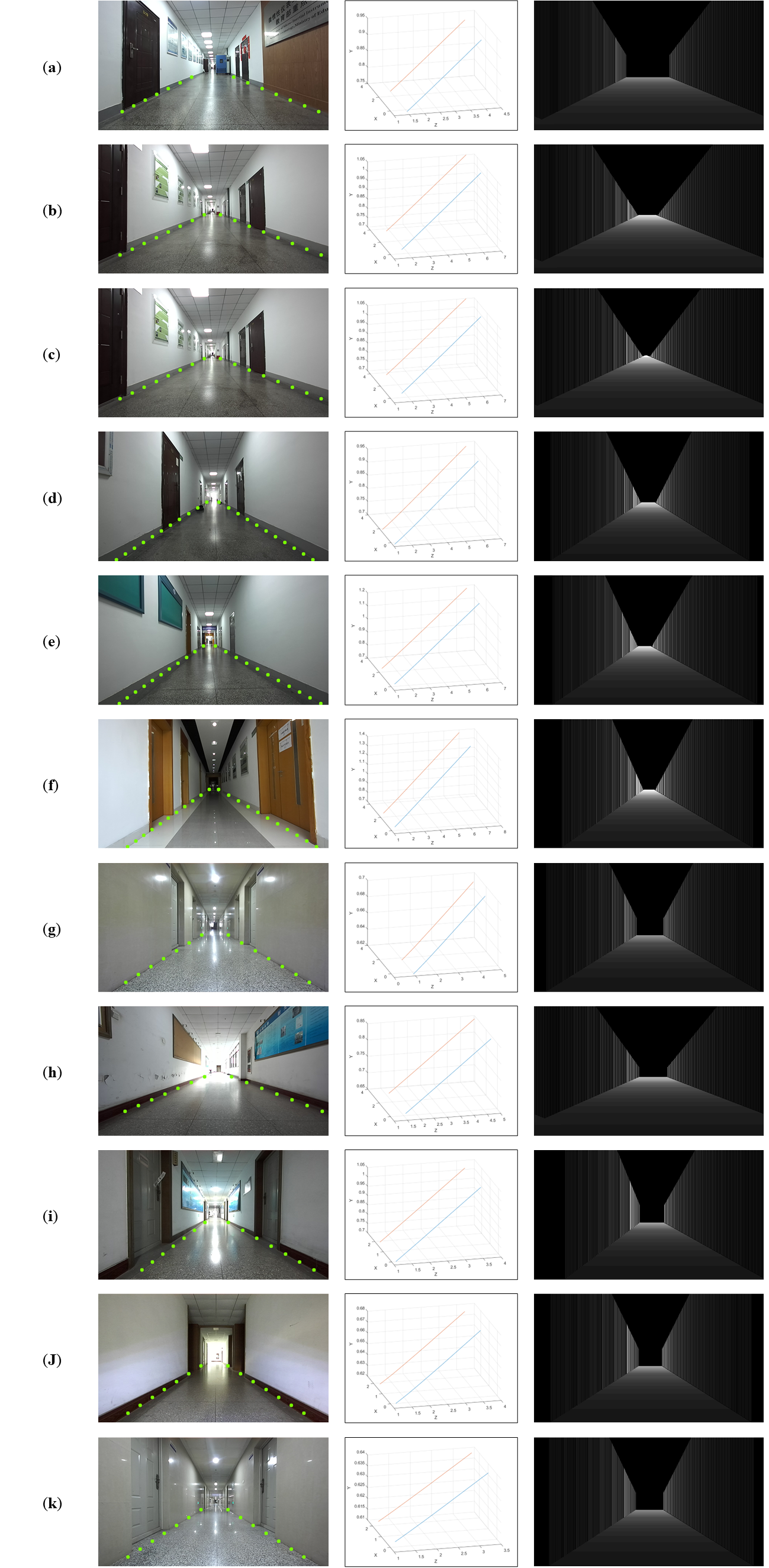}
	\caption{
		(\textbf{a})-(\textbf{k}) are estimation results in several sences. For each line, from left to right, there are edge extraction results, 3D edge estimation results and depth estimation results, respectively	.\label{fig7}}	
	\end{center}
\end{figure}

Although the method work well in the experimental scenarios, there are still some limitations:

%
%
\begin{enumerate}
	\item	Due to the use of more scene prior information, the use scenarios of the proposed method are limited. It is more suitable for corridors with characteristics of closed and straight;
	\item	Due to the ground plane assumption, the performance of the method decreases when a small slope exists in the corridor, although this scenario is rare in reality;
	\item	Because of perspective geometry, the accuracy is low at farther distances, where there is a cumulative drift in height and transversality, as shown in $~\ref{fig7}$. And the drift needs to be compensated empirically.
\end{enumerate} 

We are trying to replace straight lines with curve detection and simply model the height of the ground to tackle the above limitations, so that the proposed method can also perform well in scenes with a curvature.

The proposed algorithm can be used as an auxiliary module in the autonomous navigation and positioning system of Unmanned Aerial Vehicle(UAV) and Unmanned Ground Vehicle(UGV), such as the Simultaneous Localization and Mapping(SLAM) system,  and other applications of monocular camera systems. There are several possible applications as we can thought:

%
%

\begin{enumerate}
	\item	Inner corridor UAV safe flying;
	\item	3D positioning of the monitor camera in the corridor;
	\item	SLAM of delivery robots in the corridor.
\end{enumerate} 

In the future, we will try to apply this method to the above applications effectively. In addition, a small deep neural network for segmentation and classification can be trained to combine with this method to solve the problem of limited scenarios. Finally, the theory of our method is suitable for unsupervised training of depth estimation models.

\unskip

\bibliography{references}
\bibliographystyle{IEEEtran}

\end{document}